\documentclass[10pt,twocolumn,letterpaper]{article}

\usepackage{iccv}
\usepackage{bm}
\usepackage{times}
\usepackage{epsfig}
\usepackage{graphicx}
\usepackage{amsmath}
\usepackage{amssymb}
\usepackage{mathrsfs}

\usepackage{subcaption}

\usepackage{array, booktabs, makecell}

\usepackage[pagebackref=true,breaklinks=true,letterpaper=true,colorlinks,bookmarks=false]{hyperref}

\iccvfinalcopy 

\def\iccvPaperID{0000} 

\ificcvfinal\fi
\begin{document}

\title{Semi-supervised Fisher vector network}

\author{Petar Palasek, Ioannis Patras\\
School of Electrical Engineering and Computer Science\\
Queen Mary University of London\\
London E1 4NS, United Kingdom\\
{\tt\small p.palasek@qmul.ac.uk},  {\tt\small i.patras@qmul.ac.uk}
}

\maketitle

\begin{abstract}
In this work we explore how the architecture proposed in \cite{palasek2017discriminative}, which expresses the processing steps of the classical Fisher vector pipeline approaches, i.e.\ dimensionality reduction by principal component analysis (PCA) projection, Gaussian mixture model (GMM) and Fisher vector descriptor extraction as network layers, can be modified into a hybrid network that combines the benefits of both unsupervised and supervised training methods, resulting in a model that learns a semi-supervised Fisher vector descriptor of the input data. We evaluate the proposed model at image classification and action recognition problems and show how the model's classification performance improves as the amount of unlabeled data increases during training.
\end{abstract}

\section{Introduction}

The work of \cite{palasek2017discriminative} has shown how the standard Fisher vector pipeline for classification consisting of a local feature extraction step, a PCA dimensionality reduction step, a step where a Gaussian mixture model is learned and the final step of training a classifier can all be viewed as a single neural network, allowing finetuning of the model in a supervised, end-to-end fashion. However, even when all the mentioned steps were defined as network layers, the finetuning could start only after the layers had been initialized with parameters learned using offline batch learning procedures that require access to the whole training set at once. That is, the initialization consisted of running PCA on the whole training set, followed by using expectation-maximization to fit to data of lower dimensionality with a GMM and then training an SVM using Fisher vectors extracted using the GMM. After that the network was ready for finetuning.

What we are interested in exploring in this work are methods that would give us a way of training all the parameters in the Fisher vector pipeline in an online fashion, this way becoming suitable for applications where the training data is not available in advance, but is arriving in batches. Defining such a procedure would allow us to optimize our network using mini-batch gradient descent based methods, the usual methods used for training deep learning architectures. As the PCA dimensionality reduction step is optional in the Fisher vector pipeline, we will ignore it and first focus on defining a method that learns a Gaussian mixture model incrementally.

The final and the main goal of this work is to propose a method which allows incorporating unlabeled data into the process of the Fisher vector pipeline training, i.e.\ a method for learning a semi-supervised Fisher vector encoding of the input data. To test how the proposed method works on real data, we choose to apply it at an image classification problem; classifying tiny RGB images from the Cifar-10 \cite{krizhevsky2009learning} dataset into 10 classes, while artificially varying the amount of available labeled and unlabeled data. We also show how the same method can be used in an action recognition problem by running similar experiments on the UCF-101 dataset.

The main contributions of this work can be summarized as follows:
\begin{itemize}
    \item We describe a method for fitting Gaussian mixture models that can be used when the training data is not available all at once, but it is arriving in mini-batches, i.e.\ the data is arriving in small subsets of the training set. The same method can be used for training of the GMM defined as a network layer, described in \cite{palasek2017discriminative}. Also, the method can be run on a GPU, leading to smaller training times compared to methods that can only run on CPUs.

    \item We define a method that leverages the availability of unlabeled data for training of an improved, semi-supervised version of the Fisher vector encoding. Similarly as in \cite{palasek2017discriminative}, we define a network that extracts the Fisher vector descriptor of the input data, which we name the semi-supervised Fisher vector network. We perform a number of experiments which show how increasing the amount of unlabeled data helps improve the classification performance of our model at the problem of image classification (on CIFAR-10) and action recognition (on UCF-101).
\end{itemize}
We start by describing the problem of Gaussian mixture model fitting.

\section{Gaussian mixture model training}
We are faced with a problem of estimating a probability distribution of a random variable $X$ for which we assume that it depends on another random variable $Z$, whose values we cannot observe, i.e.\ $Z$ is a hidden random variable. If we parameterize the joint distribution of the two variables with a set of parameters ${\bm{\theta}}$, the distribution of $X$ can then be written by marginalizing over $Z$ as
\begin{equation}
p(\bm{x}|{\bm{\theta}}) = \sum_{z}p(\bm{x}, z | {\bm{\theta}}).
\end{equation}

Finding the maximum likelihood estimate parameters of the model above, that is finding the ${\bm{\theta}}$ which maximizes $L({\bm{\theta}}) = \log p(\bm{x} | {\bm{\theta}})$, 
cannot be done as a direct maximum likelihood estimation by setting the derivatives $\frac{\delta}{\delta \theta}L({\bm{\theta}})$ to zero, as this leads to entangled equations without a closed form solution. However, the parameters can be found in the case when the posterior distribution $p(z|\bm{x},{\bm{\theta}})$ is known. This observation lead to the idea of the expectation-maximization (EM) algorithm \cite{dempster1977maximum} which offers a way of finding maximum likelihood parameters using an iterative approach.

Starting from some initial estimate of the parameters ${\bm{\theta}}_{0}$, the EM algorithm works by iteratively performing two steps at each time step $t$; the E-step which determines the posterior distribution $\tilde{p_t}(z)=p(z|\bm{x},{\bm{\theta}}_{t-1})$ using the parameter estimates from the previous time step ${\bm{\theta}}_{t-1}$, and the M-step which finds new estimates of the parameters ${\bm{\theta}}_t$ as the ones that maximize the expected value of the log likelihood $L({\bm{\theta}})$ under the estimated posterior distribution from the E-step, $\tilde{p_t}$. That is, ${\bm{\theta}}_t$ is set to the ${\bm{\theta}}$ that maximizes $E_{\tilde{p_t}}\left[\log  p(\bm{x},z|{\bm{\theta}}) \right]$. This procedure is guaranteed to never worsen the log likelihood of the data \cite{dempster1977maximum}.

Following the work of \cite{neal1998view}, we can define a function $F(\tilde{p}, {\bm{\theta}})$ whose value is maximized or at least increased by both the E-step and the M-step of the EM algorithm as
\begin{equation}
\label{eq:em_f}
F(\tilde{p},{\bm{\theta}}) = E_{\tilde{p}}\left[\log  p(\bm{x},z|{\bm{\theta}}) \right] + H(\tilde{p}),
\end{equation}
with
\begin{equation}
H(\tilde{p})=-E_{\tilde{p}}\left[\log \tilde{p}(\bm{x})\right].
\end{equation}
This function can also be written in terms of a Kullback-Liebler divergence between $\tilde{p}(z)$ and $p_{\bm{\theta}}(z) = p(z |\bm{x}, {\bm{\theta}} )$ as:
\begin{equation}
\label{eq:em_f_kl}
F(\tilde{p},{\bm{\theta}}) = -D(\tilde{p}||p_{\bm{\theta}}) + L({\bm{\theta}}),
\end{equation}
where
\begin{equation}
D(\tilde{p}||p_{\bm{\theta}})=\sum_z \tilde{p}(z) \log \frac{\tilde{p}(z)}{p_{\bm{\theta}}(z)}
\end{equation}
is the KL divergence and $L({\bm{\theta}})$ is the log-likelihood, $L({\bm{\theta}})=\log p(\bm{x} | {\bm{\theta}})$.
It was shown in \cite{neal1998view} that the local and global maxima of $F(\tilde{p},{\bm{\theta}})$ are at the same time the local and global maxima of $L({\bm{\theta}})$, so maximizing $F(\tilde{p},{\bm{\theta}})$ also leads to the maximum likelihood estimate parameters ${\bm{\theta}}$ which we want to find.
When the objective function is defined as in Equation \ref{eq:em_f} the E-step of the EM algorithm maximizes $F(\tilde{p},{\bm{\theta}})$ with respect to $\tilde{p}$ and the M-step maximizes it with respect to ${\bm{\theta}}$.

\section{Online Gaussian mixture model training}
\label{sec:online_em}

In the case when the expression $E_{\tilde{p_t}}\left[\log  p(\bm{x},z|{\bm{\theta}}) \right]$ which we wish to optimize in the maximization step of the EM algorithm is not completely maximized, but only moved towards its maximum, the procedure will still result in improvements of the data likelihood \cite{dempster1977maximum}. Algorithms that perform the maximization step only partially are called generalized expectation-maximization (GEM) algorithms and are the type of algorithms that we are interested in for our problem.

As our goal is to have a way of training a GMM incorporated in a neural network as a layer, we want to be able to use the usual, gradient based optimization methods for learning the GMM's parameters. The approach which we will take for implementing the M-step will be to simply calculate the gradients of the function defined in Equation \ref{eq:em_f_kl} with respect to all of the GMM parameters and perform a number of gradient ascent steps to move the value of $F(\tilde{p},{\bm{\theta}})$ toward its maximum.

Here we define the Gaussian mixture model
as a sum of $K$ weighted Gaussians:
\begin{equation}
u_{\bm{\theta}}(\bm{x}) = \sum_{k=1}^{K}w_k u_k(\bm{x}), 
\end{equation}
with
\begin{equation}
u_k(\bm{x}) = \frac{1}{(2\pi)^{\frac{D}{2}}\lvert\bm{\Sigma}_k\rvert^{\frac{1}{2}}}\exp\left( -\frac{1}{2}(\bm{x}-\bm{\mu}_k)' \bm{\Sigma}_k^{-1} (\bm{x} - \bm{\mu}_k) \right)
\end{equation}
being the probability density function of a single Gaussian distribution. In order to meet the constraints that the weights $w_k$ need to be positive and sum up to one, we rewrite $w_k$ in terms of internal weights $\alpha_k$ as $w_k=\frac{\exp \left(\alpha_k\right)}{\sum_{l}^{K} \exp \left(\alpha_l\right)}$. The posteriors or responsibilities describing how likely it is that a sample $\bm{x}$ was generated by the $k$-th mixture component can be calculated as:
\begin{equation}
\label{eq:posterior_again}
\gamma(k) \equiv p(Z=z_k|\bm{x}) = \frac{w_k u_k(\bm{x})}{\sum_{l}^{K}{w_l u_l(\bm{x})}}.
\end{equation}
We represent the set of the GMM parameters as ${\bm{\theta}}=\{\alpha_k, \bm{\mu}_k, \bm{\Sigma}_k, k = 1, ..., K \}$, where $\alpha_k$ is the $k$-th internal component weight, $\bm{\mu}_k$ is its mean vector and $\bm{\Sigma}_k$ its covariance matrix. We will denote the set of GMM parameters at iteration $t$ as $\bm{\theta}_t$.

Finally, we will describe the steps of the generalized mini-batch EM algorithm which we use in this chapter. Given a mini-batch of training samples, i.e.\ only a small subset of the available training data, with $t$ denoting the iteration of the algorithm, we can summarize the algorithm as follows:
\begin{itemize}
    \item E-step: Maximize $F(\tilde{p}_t,{\bm{\theta}_{t-1}})$ with respect to $\tilde{p}_t$ by setting $\tilde{p_t}(z_k) \leftarrow \gamma_{t-1}(k)$, where $\gamma_{t-1}(k)$ is the posterior distribution over all the samples in the current mini-batch, calculated using the parameters $\bm{\theta}_{t-1}$ from the previous time step, ${t-1}$.
    \item M-step: Repeat $n_M$ times:
    \begin{itemize}
        \item Calculate the gradients of $F(\tilde{p}_t,{\bm{\theta}_{t-1}})$ with respect to each of the parameters $\theta_{t-1}$ in ${\bm{\theta}_{t-1}}$; $\nabla_{\theta_{t-1}} F(\tilde{p}_{t}, {\bm{\theta}}_{t-1})$.
        \item Update each $\theta_t$ in ${\bm{\theta}_{t}}$ using a gradient based method, e.g.\ $\theta_t \leftarrow \theta_{t-1} + \lambda \nabla_{\theta_{t-1}} F(\tilde{p}_{t},{\bm{\theta}}_{t-1})$.
    \end{itemize}
\end{itemize}
Note that the gradient based optimization method used in the M step is not limited to gradient ascent, but other methods such as Adagrad \cite{duchi2011adaptive} or RMSProp \cite{tieleman2012lecture} could be used too.

\subsection{Experiments on a 2D toy dataset}
In order to evaluate the performance of the method proposed above for online training of Gaussian mixture models, we first use an implementation of the standard batch expectation-maximization algorithm \cite{dempster1977maximum} from the scikit-learn \cite{scikit-learn} library and apply it to a toy problem in which we want to model a distribution of a set of points in two-dimensional space.

To generate the toy dataset we first randomly pick $K$ means, $\bm{\mu}_k$, and $K$ standard deviations, $\bm{\sigma}_k$, and use them to parameterize a set of $K$ normal distributions $\{\mathcal{N}(\bm{\mu}_k, \bm{\sigma}_k^2), k \in[0, K-1] \}$. We then define a multinomial distribution parameterized by $\{\pi_0, \ldots, \pi_{K-1}\}$, with $\sum_{k=0}^{K-1}{\pi_k =1}$, where each $\pi_k$ represents the probability of picking the $k$-th normal distribution for sampling. We finally generate a set of $n$ two-dimensional vectors by choosing one of the $K$ normal distributions according to the multinomial distribution and sampling from it. This results in data grouped in $K$ random clusters whose means and variances are known. By using the described procedure we can generate arbitrary many random point distributions, allowing us to compare the performance of our method to scikit-learn's implementation of the batch EM algorithm. Examples of the generated toy datasets are shown in Figures \ref{fig:toy_sub1} and  \ref{fig:toy2_sub1}.


\subsubsection{Parameter initialization}
\label{sub:init}
To be able to directly compare the two methods we run both of them from the same parameter initialization. We use the k-means++ algorithm \cite{arthur2007k}, a method used for initializing the locations of centroids for the k-means algorithm, to initialize the means $\bm{\mu}_k$ for each of the Gaussians in our mixture model. We initialize each of the $K$ component weights $w_k$ to $1/K$ and set the standard deviations $\bm{\sigma}_k$ to fixed values.

\subsubsection{Results}
\begin{figure}
\centering
\begin{subfigure}[t]{0.24\textwidth} 
  \centering
  \includegraphics[width=0.9\textwidth]{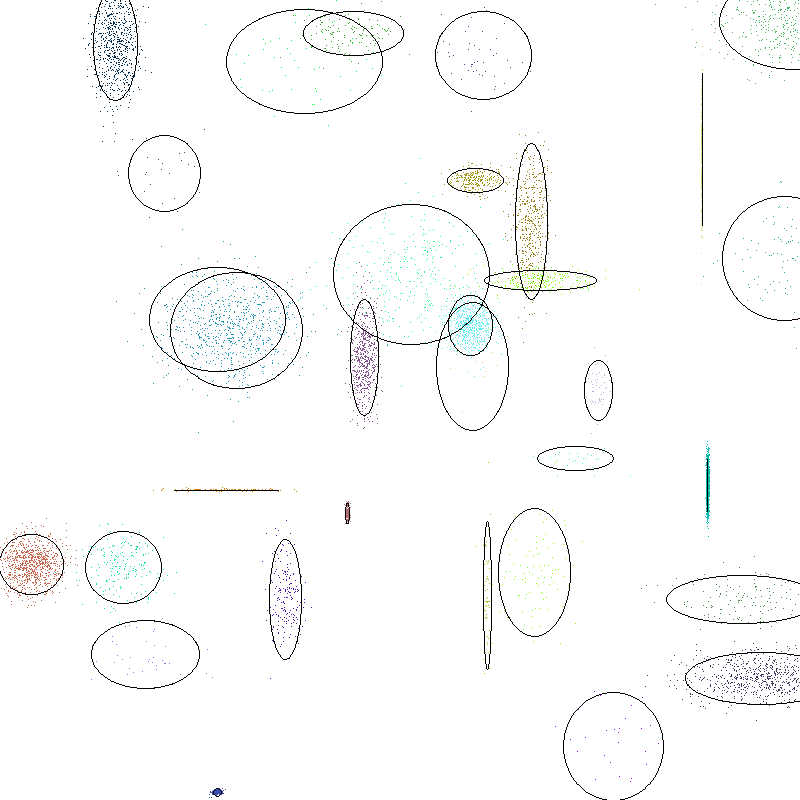}
  \caption{Groundtruth data generated by sampling from 32 Gaussians.}
  \label{fig:toy_sub1}
\end{subfigure}%
\hspace*{\fill}
\begin{subfigure}[t]{0.24\textwidth} 
  \centering
  \includegraphics[width=0.9\textwidth]{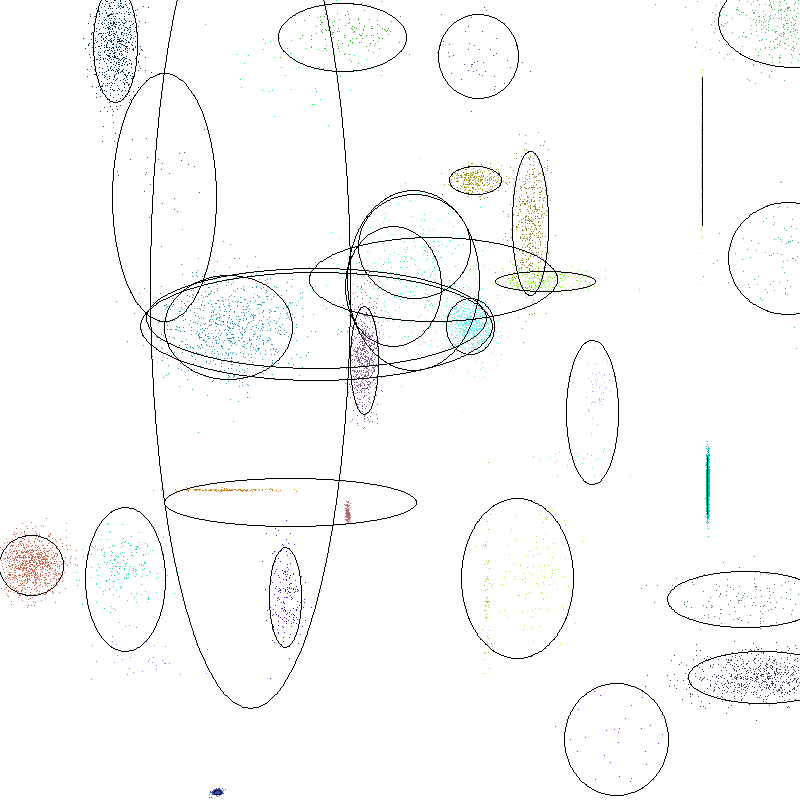}
  \caption{GMM learned using scikit-learn batch EM implementation. Mean $\log p(x_{test})=-13.48343$.}
  \label{fig:toy_sub2}
\end{subfigure}
\hspace*{\fill}
\begin{subfigure}[t]{0.24\textwidth} 
  \centering
  \includegraphics[width=0.9\textwidth]{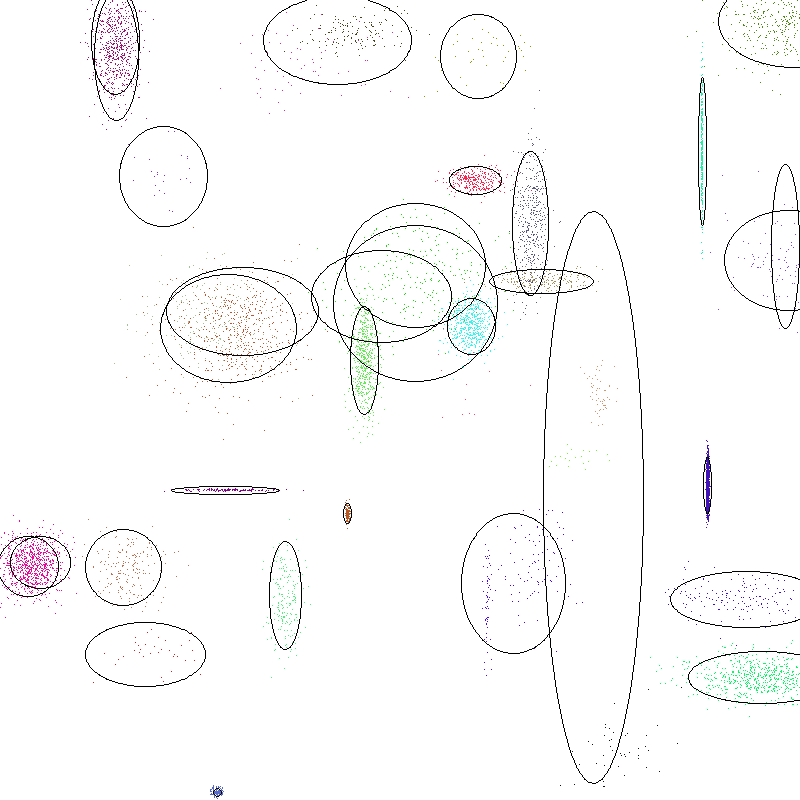}
  \caption{GMM learned using the proposed algorithm with Adagrad, 1\,000 M-step iterations, learning rate 1. Mean $\log p(x_{test})=-10.9322$.}
  \label{fig:toy_sub3}
\end{subfigure}%
\hspace*{\fill}
\begin{subfigure}[t]{0.24\textwidth} 
  \centering
  \includegraphics[width=0.9\textwidth]{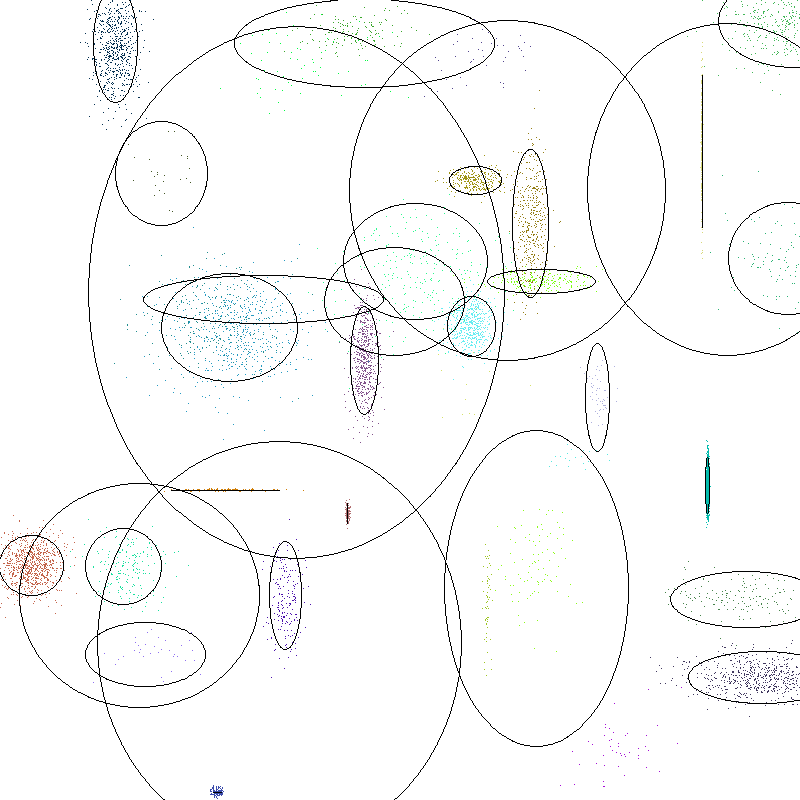}
  \caption{GMM learned using the proposed algorithm with Nesterov momentum, 1\,000 M-step iterations, learning rate 10. Mean $\log p(x_{test})=-10.9354$.}
  \label{fig:toy_sub4}
\end{subfigure}
\caption{Experiments on the 2D toy dataset. The first subfigure shows the generated dataset, the second subfigure shows the GMM learned using scikit-learn's EM implementation, the third subfigure shows the GMM learned using the proposed EM algorithm with Adagrad and the fourth subfigure shows the GMM learned using the proposed EM algorithm with Nesterov mementum. The training set contained 10\,000 samples and the mini-batch size in the last two experiments was set to 10\,000.}
\label{fig:toy}
\end{figure}

\begin{figure}
\centering
\begin{subfigure}[t]{0.24\textwidth} 
  \centering
  \includegraphics[width=0.9\textwidth]{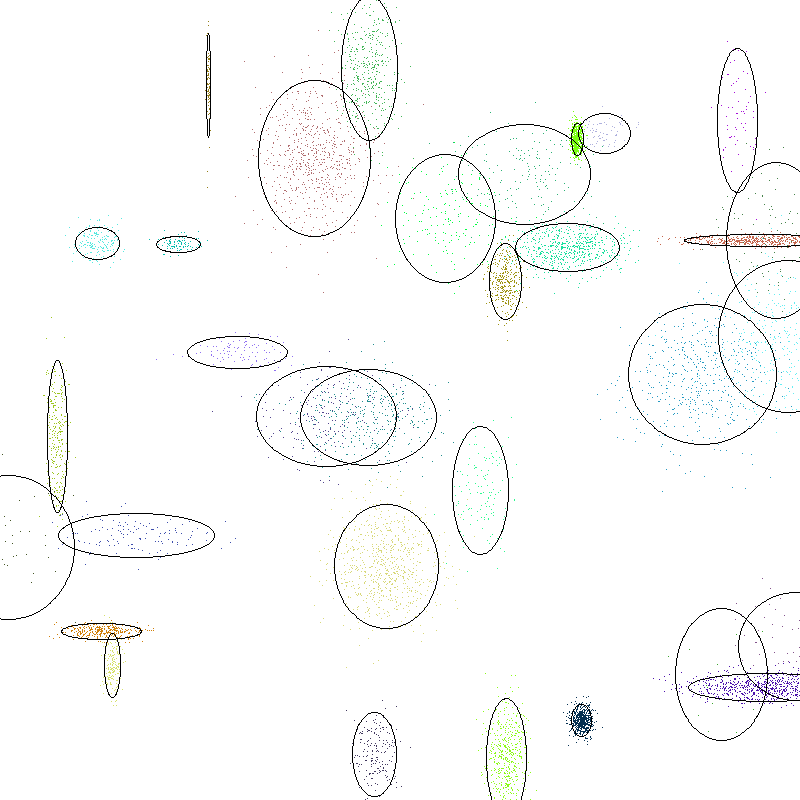}
  \caption{Groundtruth data generated by sampling from 32 Gaussians.}
  \label{fig:toy2_sub1}
\end{subfigure}%
\hspace*{\fill}
\begin{subfigure}[t]{0.24\textwidth} 
  \centering
  \includegraphics[width=0.9\textwidth]{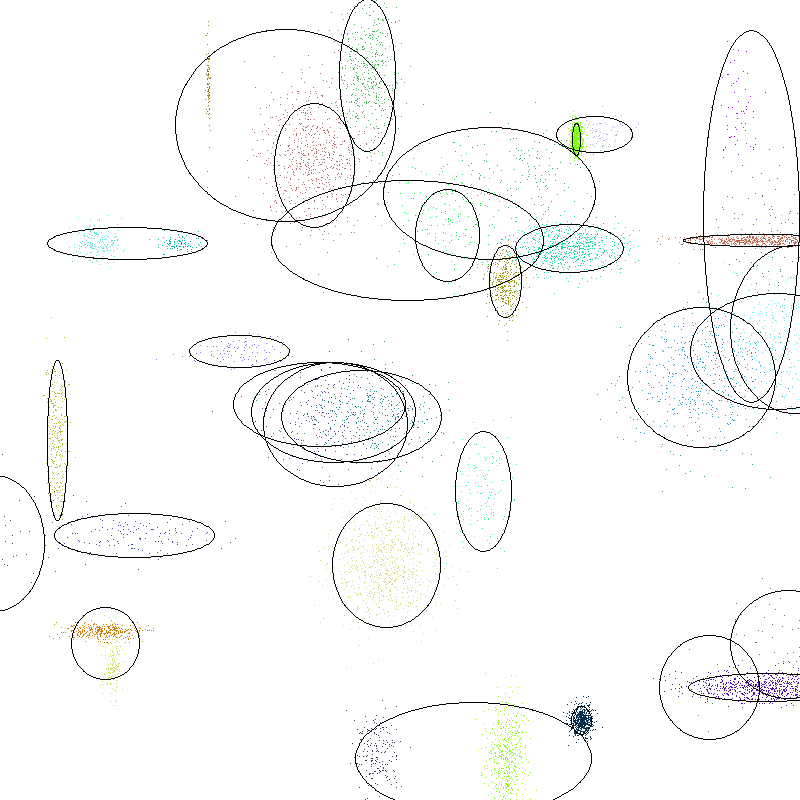}
  \caption{GMM learned using scikit-learn batch EM implementation. Mean $\log p(x_{test})=-14.46651$.}
  \label{fig:toy2_sub2}
\end{subfigure}
\hspace*{\fill}
\begin{subfigure}[t]{0.24\textwidth} 
  \centering
  \includegraphics[width=0.9\textwidth]{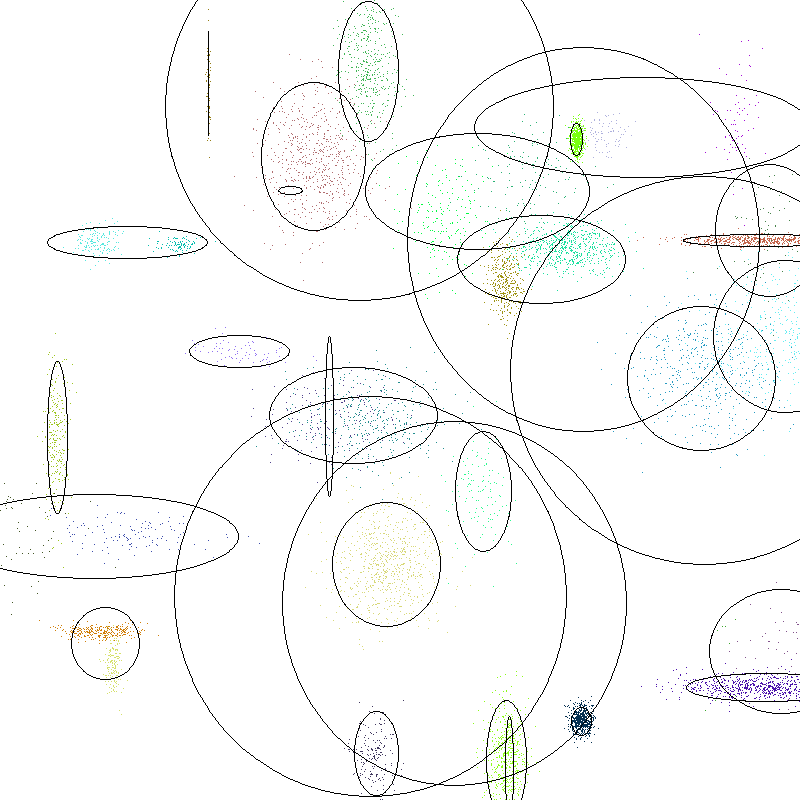}
  \caption{GMM learned using the proposed algorithm with Nesterov momentum, 1\,000 M-step iterations, learning rate 10. Mean $\log p(x_{test})=-11.76504$.}
  \label{fig:toy2_sub3}
\end{subfigure}%
\hspace*{\fill}
\begin{subfigure}[t]{0.24\textwidth} 
  \centering
  \includegraphics[width=0.9\textwidth]{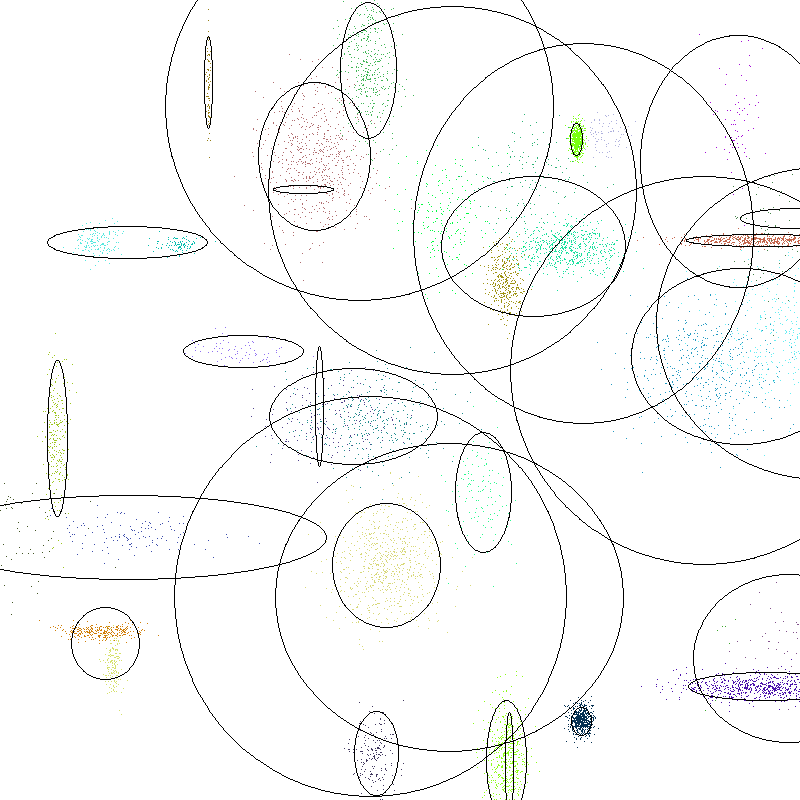}
  \caption{GMM learned using the proposed algorithm with RMSProp, 1\,000 M-step iterations, learning rate 0.01. Mean $\log p(x_{test})=-11.80651$.}
  \label{fig:toy2_sub4}
\end{subfigure}
\caption{Experiments on the 2D toy dataset. The first subfigure shows the generated dataset, the second subfigure shows the GMM learned using scikit-learn's EM implementation, the third subfigure shows the GMM learned using the proposed EM algorithm with Nesterov momentum and the fourth subfigure shows the GMM learned using the proposed EM algorithm with RMSProp. The training set contained 10\,000 samples and the mini-batch size in the last two experiments was set to 10\,000.}
\label{fig:toy2}
\end{figure}

\begin{figure}
\centering
\begin{subfigure}[t]{0.24\textwidth} 
  \centering
  \includegraphics[width=0.9\textwidth]{groundtruth_data_image_30042027.png}
  \caption{Groundtruth data generated by sampling from 32 Gaussians.}
  \label{fig:toy3_sub1}
\end{subfigure}%
\hspace*{\fill}
\begin{subfigure}[t]{0.24\textwidth} 
  \centering
  \includegraphics[width=0.9\textwidth]{sklearn_result_image_30042027.png}
  \caption{GMM learned using scikit-learn batch EM implementation. Mean $\log p(x_{test})=-13.48343$.}
  \label{fig:toy3_sub2}
\end{subfigure}
\hspace*{\fill}
\begin{subfigure}[t]{0.24\textwidth} 
  \centering
  \includegraphics[width=0.9\textwidth]{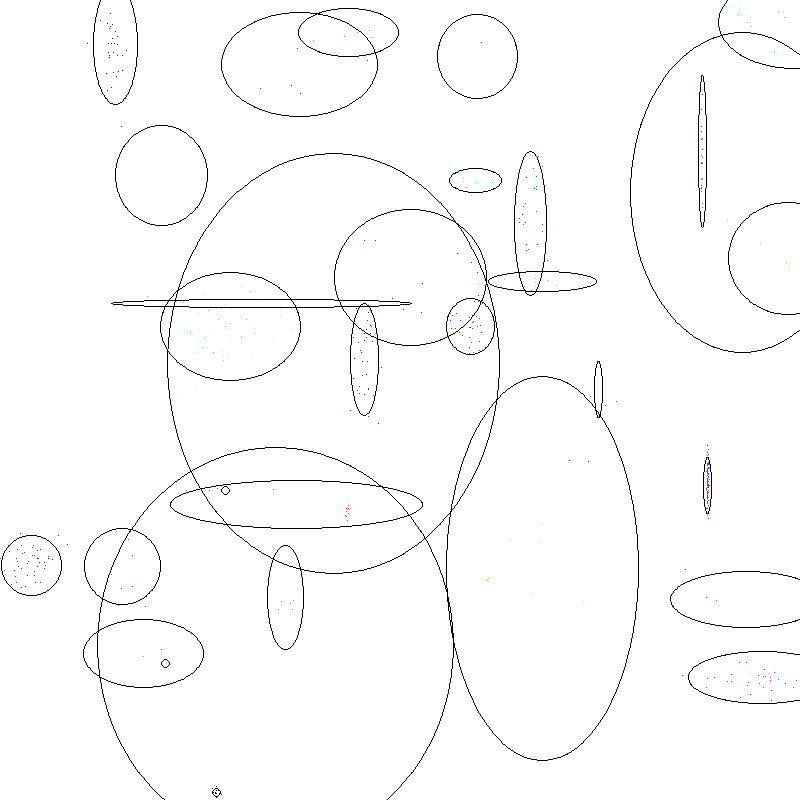}
  \caption{GMM learned using the proposed algorithm with Adagrad, 100 M-step iterations, learning rate 1. $\log p(x_{test})=-11.03016$.}
  \label{fig:toy3_sub3}
\end{subfigure}%
\hspace*{\fill}
\begin{subfigure}[t]{0.24\textwidth} 
  \centering
  \includegraphics[width=0.9\textwidth]{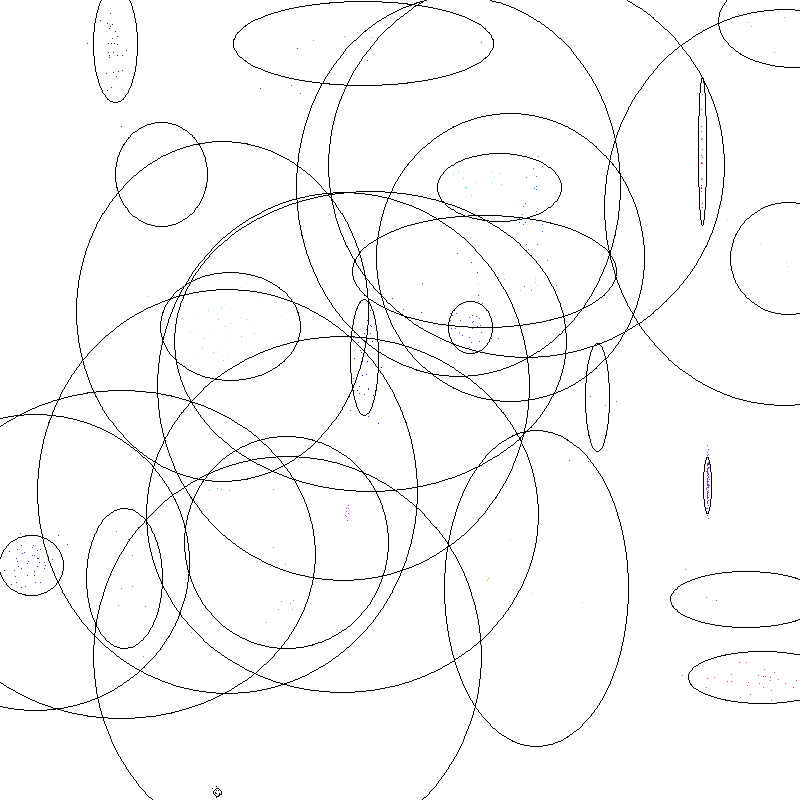}
  \caption{GMM learned using the proposed algorithm with Adagrad, 10 M-step iterations, learning rate 1. $\log p(x_{test})=-11.14458$.}
  \label{fig:toy3_sub4}
\end{subfigure}
\caption{Experiments on the 2D toy dataset. The first subfigure shows the generated dataset, the second subfigure shows the GMM learned using scikit-learn's EM implementation, the third and the fourth subfigures show the GMM learned using the proposed EM algorithm with Adagrad, but with different numbers of M-step iterations. The training set contained 10\,000 samples and the mini-batch size in the last two experiments was set to 500.}
\label{fig:toy3}
\end{figure}

We generated a number of random 2D toy datasets consisting of 10\,000 training and 10\,000 testing samples from 32 different Gaussian distributions. We show the generated datasets in the top left corners of Figures \ref{fig:toy}, \ref{fig:toy2} and \ref{fig:toy3}. We then use the implementation of the batch EM algorithm from the scikit-learn library to fit a GMM with 32 components, starting from the intialization described above. The training was stopped after the improvement of the training data log likelihood between two iterations fell below a certain threshold value. 

The results acquired using the batch EM algorithm are shown in top right corners of Figures \ref{fig:toy}, \ref{fig:toy2} and \ref{fig:toy3}. After that we run the version of the EM algorithm we described in Section \ref{sec:online_em} using different gradient based optimization methods, different learning rates and different number of M-step iterations. We run the training for a fixed number of iterations. In the first two experiments the number of samples given in a mini-batch was the same as the total number of available training samples. The results of running the proposed method are shown in the bottom row of Figures \ref{fig:toy}, \ref{fig:toy2}. In the last experiment we use a mini-batch size of 500 and show the results in the bottom row of Figure \ref{fig:toy3}. We also report the log likelihood of the testing data in each of the experiments below the figures. It can be seen that the learned Gaussians identified most of the groundtruth clusters.  
The proposed algorithm also learned some Gaussians with a small variance, which also increased the average likelihood. The values of the variances during training were limited to not fall bellow a predefined value.

\section{Semi-supervised Fisher vector encoding}
\label{sec:semisup}

\begin{figure*}[t]
\begin{center}
\includegraphics[width=0.8\linewidth]{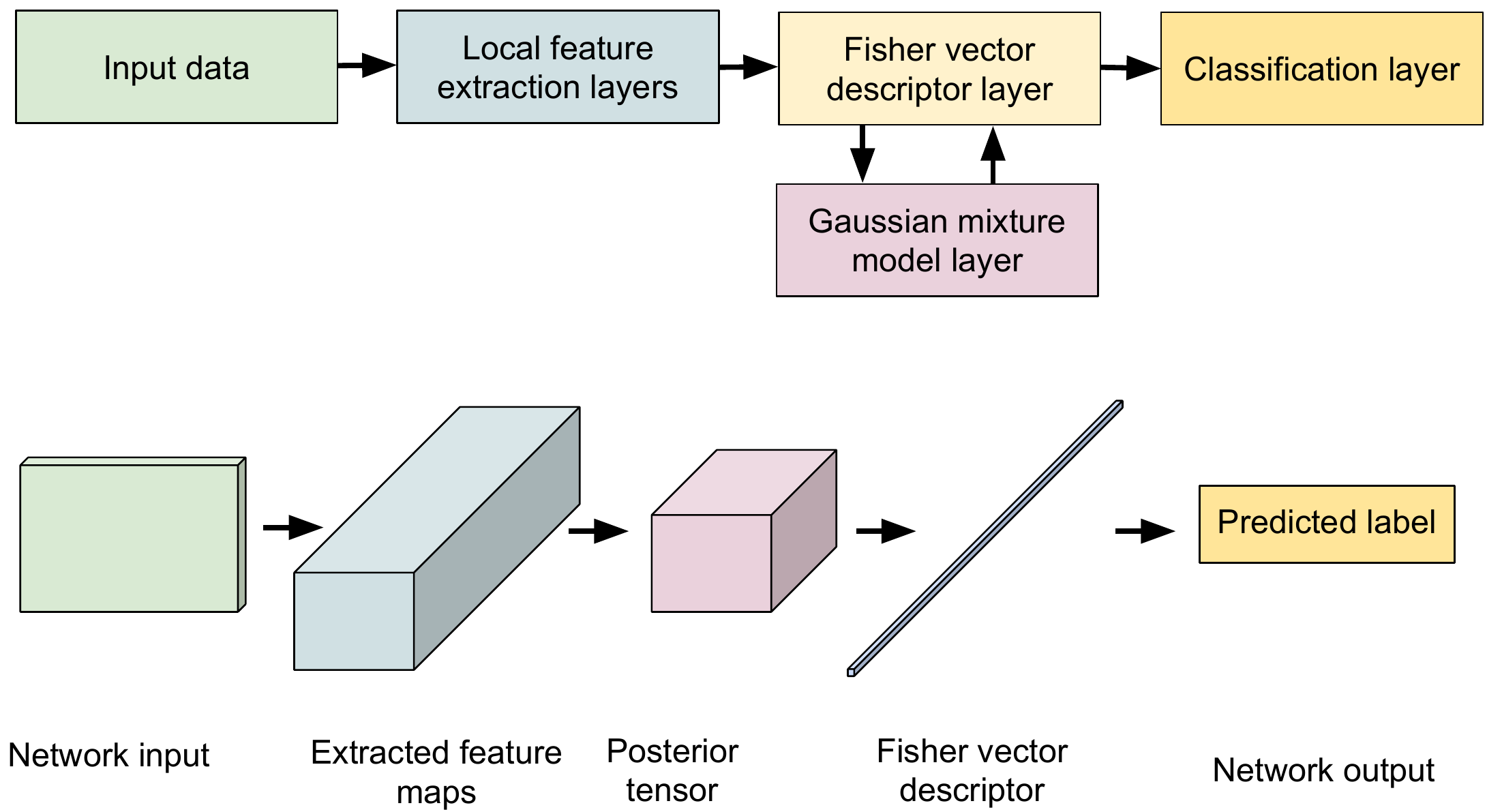}

\end{center}
   \caption{An illustration of the proposed architecture. It is a simplified version of the network proposed in 
   \cite{palasek2017discriminative}, without the spatio-temporal pooling layer and the dimensionality reduction layer. For example, we can think of the network input as an RGB image that is fed through local feature extraction layers that give some feature maps as output. The "pixels" of the feature maps are modeled with a GMM layer that outputs a tensor of posteriors for each "pixel" in the given feature map. Feeding the extracted feature maps and the posterior tensor into the Fisher vector descriptor layer gives the FV descriptor as an output. The FV descriptor is then fed into the classification layer that outputs a class prediction for the image given at the input of the network.}

\label{fig:architecture2}
\end{figure*}

In this section we will define a method for leveraging unlabeled data in the Fisher vector classification pipeline. An illustration of the architecture that we describe in this chapter is shown in Figure \ref{fig:architecture2}.

Having a method for training Gaussian mixture models with a gradient based optimization method which we described in the previous section (Section \ref{sec:online_em}) gives us a natural way to build a hybrid network that is trained with the goal of optimizing two different objective functions; an unsupervised objective function and a supervised objective function. The unsupervised objective function is the one whose goal is to fit a Gaussian mixture model to the given data and the supervised objective function would be the one that forces the network to output the correct label given a training sample. We already have both of the needed functions defined - the unsupervised objective function, $C_{unsup}$ is $F(\tilde{p},{\bm{\theta}})$ from Equation \ref{eq:em_f} and the supervised objective function, which we will denote as $C_{sup}$, is the squared hinge loss 
also used in \cite{palasek2017discriminative}:
\begin{equation}
\label{eq:unsup}
C_{unsup} \equiv F(\bm{x}, \tilde{p},{\bm{\theta}}) = -\sum_z \tilde{p}(z) \log \frac{\tilde{p}(z)}{p_{\bm{\theta}}(z)} + \log p(\bm{x} | {\bm{\theta}}),
\end{equation}
and
\begin{equation}
\label{eq:sup}
C_{sup} \equiv C(\bm{x}, \bm{y}, \bm{w}, \bm{b}) = \frac{\lambda}{2}||\bm{w}||^2 + \sum_j^m \max \left( 0, 1 - y_j \cdot s_j \right)^2,
\end{equation}
where $\bm{s}=\bm{x}\bm{w}^T+\bm{b}$ is the SVM score, $s_j$ denotes the $j$-th element of $\bm{s}$ and $\bm{y}$ is the label $l$ of sample $\bm{x}$ encoded as a vector where, in case of dealing with $m$ classes, $m-1$ elements are set to -1 and a single element at position $l$ is set to 1. 
After this, we can define the combined objective cost as
\begin{equation}
C_{hybrid} = C_{sup} - \lambda \cdot C_{unsup},
\end{equation}
where $\lambda$ is a weighting coefficient. We subtract the unsupervised cost in order to turn its optimization from a maximization into a minimization problem.

Trying to optimize the local feature extraction layers' weights jointly with the GMM and SVM layers is not straightforward as the feature extraction layers' weights might collapse to zeros, leading to a trivial but not meaningful solution that minimizes the unsupervised part of the cost. Here we focus only at the GMM and SVM layers and keep the parameters of the local feature extraction layers fixed. The training of the proposed model can in this case be done as follows.

We loop through the training set consisting of both labeled and unlabeled data, at each step taking a mini-batch from the set. Each chosen mini-batch can either consist of a mix of labeled and unlabeled samples, all the samples can be unlabeled or all the samples can be labeled. For each mini-batch we first perform the E-step of the EM algorithm described in Section \ref{sec:online_em}, setting $\tilde{p}$ to the one that minimizes $-\lambda \cdot C_{unsup}$ while keeping the other parameters fixed. This is followed by $n_M$ M-steps in which all the GMM parameters $\theta$ from $\bm{\theta}$ are updated using a gradient-based optimization method. Finally, we update all the parameters of both the GMM and the SVM layers by taking gradients of the combined cost, $C_{hybrid}$, with respect to each of the parameters and moving towards its minimum again using a gradient-based optimization method. In the case when a sample is unlabeled, its supervised part of the cost is set to 0. We evaluate the proposed method in the following section.

\section{Experiments and results}

In this section we report how we evaluated the proposed semi-supervised Fisher vector network on the problems of image classification and action recognition.

\subsection{Image classification on CIFAR-10}

The CIFAR-10 dataset \cite{krizhevsky2009learning} is a standard image classification dataset containing small $32\times32$ px images from 10 different categories.
In our experiments the images are preprocessed by subtracting the mean calculated on the training set from each image.

The architecture that we will use for the experiments we run on the CIFAR-10 dataset is shown in Figure \ref{fig:architecture2}, where the local feature extraction layers were replaced by the layers taken from the VGG-16 network \cite{simonyan2014very}, pretrained on the ImageNet dataset.
In each experiment we will specify which layer is used as the input layer to the GMM layer.

\subsubsection{Architecture without a GMM layer}
The first experiments that we run are with an architecture that does not contain a GMM layer, that is, the outputs of the VGG-16 network layer are fed straight into the SVM layer. First we keep the VGG layer weights fixed and only train the SVM layer for a fixed number of epochs. We then try training the whole network using the SVM cost (Equation \ref{eq:sup}). We check how adding an additional convolutional layer with 512 $1\times1$ px filters affects the accuracy. We also try adding an additional fully connected layer with 16416 units. We use different layers of VGG-16 as the input to the SVM layer. The results are shown in Table \ref{table:ex1}.

\begin{table*}[htb]
\centering
\caption{Classification results on the CIFAR-10 test set using different outputs of the VGG-16 network directly fed into the SVM layer. We train the network using RMSProp for 100 epochs.}
\scalebox{0.96}
{
\begin{tabular}{lccc}
\textbf{Output layer}  & \textbf{Output shape} & \textbf{Parameters trained}   & \textbf{Accuracy}          \\
\hline
conv\_3\_3  & (128, 256, 8, 8) & SVM & 69.96\%\\
conv\_3\_3  & (128, 256, 8, 8) & all & 81.97\%\\
conv\_3\_3 + conv layer & (128, 32768) & all& 81.99\%\\
conv\_3\_3 + FC layer & (128, 16416) & all & 82.33\%\\
conv\_4\_3  & (128, 512, 4, 4) & all & 85.44\%\\
conv\_5\_3  & (128, 512, 2, 2) & all & 86.39\%\\
\end{tabular}
}
\label{table:ex1}
\end{table*}

\subsubsection{Architecture with a GMM layer trained offline}

In these experiments we pass images from the training set through the VGG network and model the outputs of its different layers with a GMM trained using an implementation of the EM algorithm from the scikit-learn library. Once the GMM parameters have been learned, we extract Fisher vector representations of the input images and use them to train a classifier by optimizing the SVM squared hinge loss. We use mini-batch gradient descent with momentum to do the training for 100 epochs, with the learning rate set to $0.001$, momentum $0.9$ and mini-batch size of 128. The number of GMM components was 64.

\begin{table*}[htb]
\centering
\caption{Classification results on the CIFAR-10 test set using different outputs of the VGG-16 network. The GMM layer contained 64 components and was trained using scikit-learn's implementation of the batch EM algorithm. The SVM was trained using gradient descent with momentum for 100 epochs.}
\scalebox{0.96}
{
\begin{tabular}{lcccc}
\textbf{Output layer}  & \textbf{Output shape}   & \textbf{FV size} & \textbf{Parameters trained} & \textbf{Accuracy}          \\
\hline
pool\_5 & (128, 512, 1, 1) & (128, 65600) & SVM & 53.79\%\\
conv\_5\_3 & (128, 512, 2, 2) & (128, 65600) & SVM & 57.10\%\\
conv\_4\_3  & (128, 512, 4, 4) & (128, 65600) & SVM & 72.48\%\\
conv\_3\_3  & (128, 256, 8, 8) & (128, 32832) & SVM & 77.73\%\\
conv\_3\_3  & (128, 256, 8, 8) & (128, 32832) & all & 86.03\%\\
\end{tabular}
}
\label{table:ex2}
\end{table*}

\subsubsection{Architecture with a GMM layer trained using the proposed hybrid objective function}

In these experiments we will evaluate the proposed method for training a semi-supervised Fisher vector network, as described in Section \ref{sec:semisup}. We are given a training set consisting of a fixed number of labeled samples and we vary the number of unlabeled data throughout the experiments. We start by initializing the GMM layer using the k-means++ algorithm \cite{arthur2007k}, as it was done in Section \ref{sub:init}. Only the labeled data is used to do the initialization. 

The optimization method we use for the M-step is Adagrad with the initial learning rate set to 1. The updates with respect to the supervised objective function are done using RMSProp using an initial learning rate that we mention for each of the experiments.

\begin{figure*}[htb]
\includegraphics[width=1\textwidth]{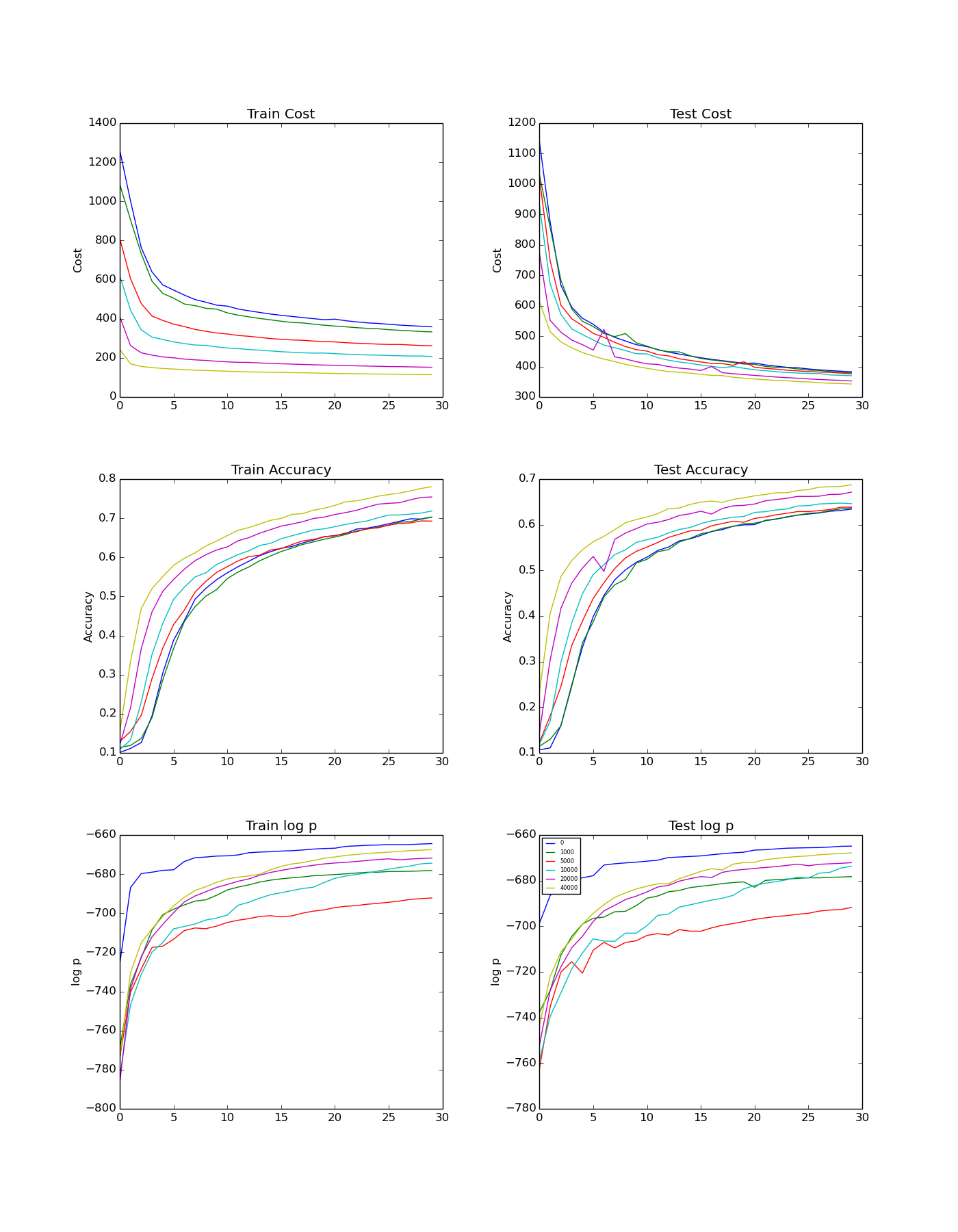}
\caption{Training of the semi-supervised Fisher vector network on CIFAR-10 using 10\,000 labeled samples and changing the number of unlabeled samples. The training was run for 30 epochs, the learning rate was 0.0001 and we used 5 M-step iterations.}
\label{fig:exp1}
\end{figure*}

In the first experiment on the CIFAR-10 dataset we fix the number of labeled samples to 10\,000 and change the amount of unlabeled samples from 0, 1\,000, 5\,000, 10\,000, 20\,000 to 40\,000. The learning rate is set to 0.0001 and we perform 5 M-step iterations for each mini-batch. The training charts for this experiment can be seen in Figure \ref{fig:exp1}.

\begin{figure*}[htb]
\includegraphics[width=1\textwidth]{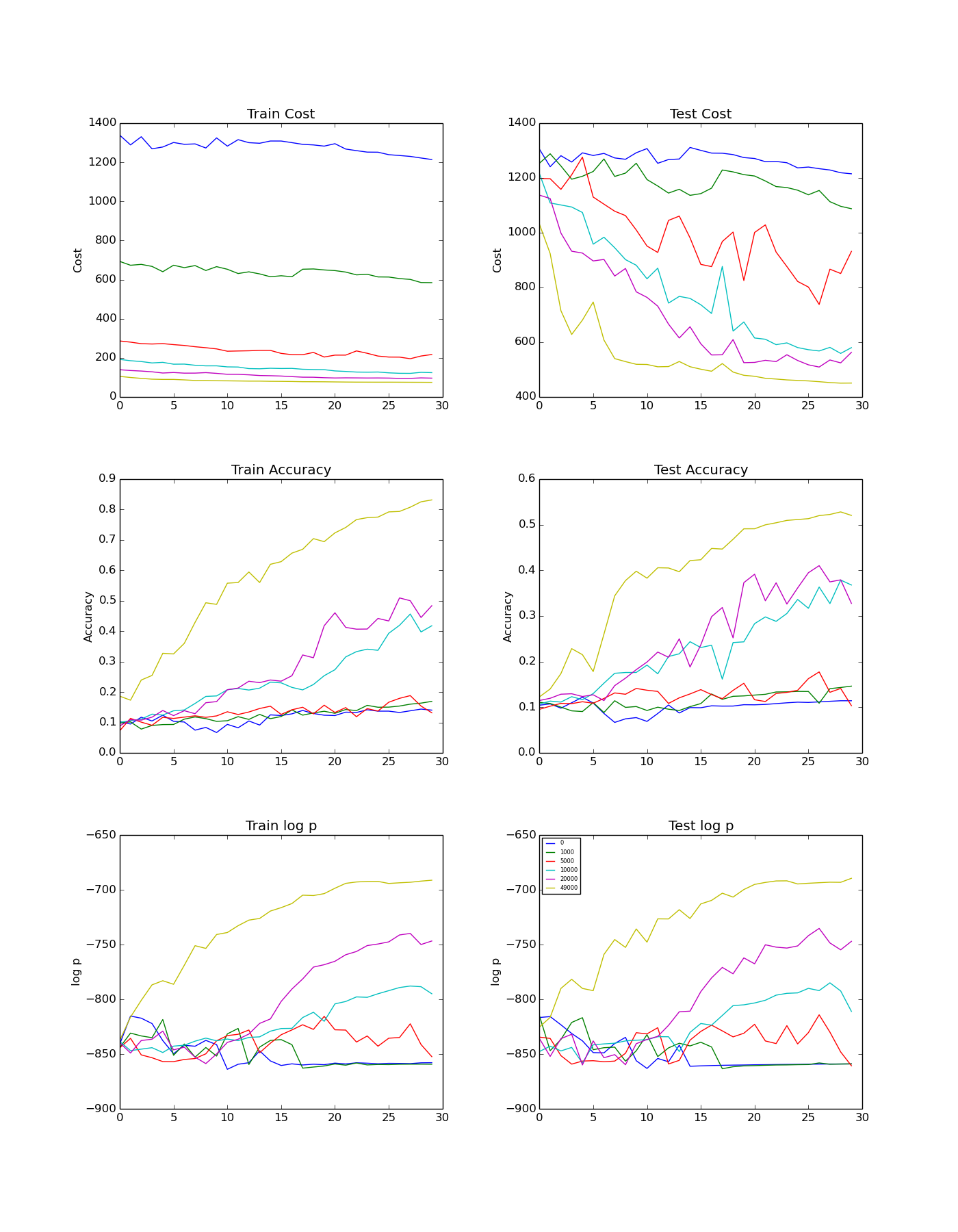}
\caption{Training of the semi-supervised Fisher vector network on CIFAR-10 using 1\,000 labeled samples and changing the number of unlabeled samples. The training was run for 30 epochs, the learning rate 0.0001 and we used only a single M-step iteration.}
\label{fig:exp2}
\end{figure*}

In the second experiment on the CIFAR-10 dataset we fix the number of labeled samples to 1\,000 and change the amount of unlabeled samples from 0, 1\,000, 5\,000, 10\,000, 20\,000 to 49\,000. The learning rate is set to 0.0001 and we perform only a single M-step iteration for each mini-batch. The training charts for this experiment can be seen in Figure \ref{fig:exp2}. In all of the mentioned experiments the coefficient $\lambda$ was set to 0.1.


\subsection{Action recognition on UCF-101}

In order to evaluate how the semi-supervised Fisher vector network performs on an action recognition problem when the amount of unlabeled data is being increased we do the following experiments. We use the representation used in \cite{palasek2016marmi}, where instead of performing dense sampling as in \cite{palasek2017discriminative}, only random subvolumes are extracted from a video for calculating the FV. We
use the layer conv5\_3 from the VGG-16 network as the feature extraction layer. We extract 1\,000 subvolumes from each video and pool them spatially and temporally as illustrated in the mentioned figure. As we are not interested in learning the PCA dimensionality reduction mapping, we use the PCA learned in the experiments from 
\cite{palasek2017discriminative}
to lower the dimensionality of each extracted subvolume to 100. We then end up with 1\,000 feature vectors of size 100 for each video, which is what we feed into our GMM layer. The rest of the training process is the same as with the experiments we performed on the CIFAR-10 dataset. We also start by initializing the GMM using the k-means++ algorithm on the labeled part of the training data. The optimization method used in the M-step is Adagrad with the initial learning rate set to 1, and the optimization method used to update the parameters with respect to the supervised objective function is RMSProp, as it was in the CIFAR-10 experiments.



\begin{figure*}[htb]
\includegraphics[width=1\textwidth]{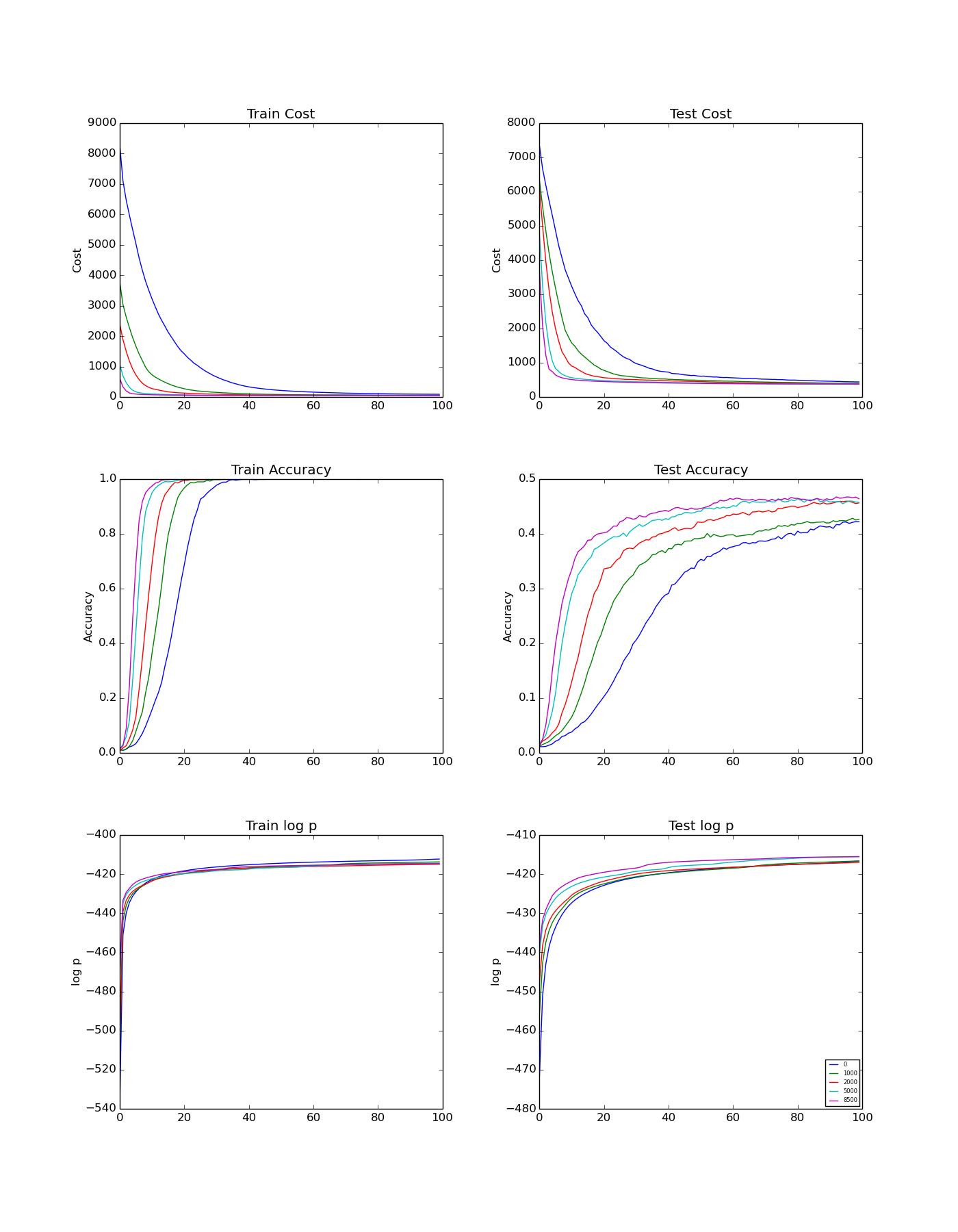}
\caption{Training of the semi-supervised Fisher vector network on UCF-101, 64 GMM components, using 1000 labeled samples and changing the number of unlabeled data. Learning rate 0.001, 5 M-steps, $\lambda=0.1$.}
\label{fig:ucf_2}
\end{figure*}

We start by using 1\,000 labeled samples and a varying amount of unlabeled samples (0, 1\,000, 5\,000 and 8\,500) to train the semi-supervised Fisher vector network with 64, 128 and 256 GMM components using the proposed method for 100 epochs. We show the effect of using different amount of unlabeled data for the network with 64 GMM components in \ref{fig:ucf_2}. Very similar behavior was observed when 128 or 256 GMM components were used, so we do not include the corresponding figures. In all of the mentioned experiments the coefficient $\lambda$ was set to 0.1. We show the achieved classification performance from these experiments in Table \ref{table:ucf_gmm}.

\begin{table}[htb]
\centering
\caption{Classification results on the UCF-101 test set using semi-supervised Fisher vector networks with different numbers of GMM components. The number of labeled samples was fixed to 1\,000, $\lambda=0.1$ and the training ran for 100 epochs.}
\scalebox{0.96}
{
\begin{tabular}{lccc}
\thead{\textbf{Number unlabeled}\\ \textbf{samples}}  & \thead{\textbf{64 GMM}\\ \textbf{components}}   & \thead{\textbf{128 GMM}\\ \textbf{components}} & \thead{\textbf{256 GMM}\\ \textbf{components}} \\
\hline
0 & 42.21\% & 41.76\% & 42.62\% \\
1\,000 & 42.65\% & 43.73\% & 43.51\% \\
2\,000 & 45.65\% & 43.78\% & 44.59\% \\
5\,000 & 45.76\% & 45.76\% & 47.03\% \\
8\,500 & 46.41\% & 46.16\% & 46.32\% \\
\end{tabular}
}
\label{table:ucf_gmm}
\end{table}

In the next experiment we chose to train the GMM component of the network offline, using the standard batch EM algorithm using both unlabeled and labeled data, and then we finetune the network using the labeled data only. We do this in order to be able to compare the achieved performance to the performance of the proposed semi-supervised training that updates all the network parameters at the same time, in an online fashion. The results shown in Table \ref{table:ucf_gmm_offline} demonstrate that the performance of the network where the GMM part was trained offline almost always surpasses the performance of the network trained in an online fashion. This is an expected result as the batch EM algorithm has access to the information from the whole training set at each step of the training. On the other hand, our online version can only make updates to the network parameters based only on the information contained in the current mini-batch.

\begin{table}[htb]
\centering
\caption{Classification results on the UCF-101 test set using semi-supervised Fisher vector networks with different numbers of GMM components. The GMMs are trained offline using labeled and unlabeled data and the networks are finetuned using the 1\,000 labeled samples. The training ran for 100 epochs.}
\scalebox{0.96}
{
\begin{tabular}{lccc}
\thead{\textbf{Number unlabeled}\\ \textbf{samples}}  & \thead{\textbf{64 GMM}\\ \textbf{components}}   & \thead{\textbf{128 GMM}\\ \textbf{components}} & \thead{\textbf{256 GMM}\\ \textbf{components}} \\
\hline
0 & 46.70\% & 46.08\% & 43.76\% \\
1\,000 & 46.78\% & 46.05\% & 44.16\% \\
2\,000 & 46.08\% & 45.67\% & 44.11\% \\
5\,000 & 47.41\% & 46.83\% & 44.64\% \\
8\,500 & 45.94\% & 46.19\% & 44.70\% \\
\end{tabular}
}
\label{table:ucf_gmm_offline}
\end{table}

In order to see how much information is gained by increasing the available number of labeled samples compared to when the number of unlabeled samples is increased, we run a new set of experiments in which we do not use any unlabeled data, we only modify the amount of available labeled data. The results of this experiment are shown in Table \ref{table:ucf_labeled_only}. When comparing to the results of the previous experiment shown in Table \ref{table:ucf_gmm} we can see that we only needed to add on average 600 labeled samples to surpass the performance of the semi-supervised Fisher vector network that was trained with 1\,000 labeled and 8\,500 unlabeled samples.

\begin{table}[htb]
\centering
\caption{Classification results on the UCF-101 test set using the semi-supervised Fisher vector network trained without any unlabeled data and different amounts of labeled data.}
\scalebox{0.96}
{
\begin{tabular}{lc}
\thead{\textbf{Number labeled}\\ \textbf{samples}}  & \thead{\textbf{Test}\\ \textbf{accuracy}} \\
\hline
1\,000 & 42.21\%  \\
1\,200 & 42.47\%  \\
1\,400 & 44.67\%  \\
1\,600 & 47.12\%  \\
\end{tabular}
}
\label{table:ucf_labeled_only}
\end{table}



In order to see how the performance of the proposed network changes as a function of the mini-batch size used during training, we run a set of experiments where we trained the network in a semi-supervised way while changing the size of the mini-batch. We trained the network using mini-batch size 1, 5, 50, 100, 250 and 500. We show the results in Table \ref{table:changing_mb_size}. It can be seen that, out of the different mini-batch sizes we have used for the experiment, the best performance is achieved using a mini-batch size of 50.

\begin{table}[htb]
\centering
\caption{Classification results on the UCF-101 test set using the semi-supervised Fisher vector network trained using different mini-batch sizes. The numbers of both labeled and unlabeled samples were fixed to 1\,000, we used 64 GMM components and the learning rate was set to 0.001.}
\scalebox{0.96}
{
\begin{tabular}{lc}
\thead{\textbf{Mini-batch}\\ \textbf{size}}  & \thead{\textbf{Test}\\ \textbf{accuracy}} \\
\hline
1 & 12.24\%  \\
5 & 37.39\%  \\
50 & 46.31\% \\
100 & 42.65\%  \\
250 & 39.44\%  \\
500 & 32.29\%  \\
\end{tabular}
}
\label{table:changing_mb_size}
\end{table}

\section{Discussion and conclusion}

In the first experiments 
(Table \ref{table:ex1}) we have shown how training an SVM using different output layers of the VGG-16 network affects the classification accuracy evaluated on the CIFAR-10 dataset. We can notice that using higher layers improved the performance as these layers capture more high-level features making it easier for the SVM to discriminate between the classes.
We have also shown that finetuning the whole network leads to significant improvements in the performance. This is expected as the VGG-16 network that we use was pretrained on a different dataset (ImageNet), so the weights are not tailored for the CIFAR-10 dataset.

In the experiments where a Fisher vector encoding was used (Table \ref{table:ex2}) we showed that having feature maps of larger spatial size results in better performance. This is because the Fisher vector combines local features into a global descriptor and having feature maps with an image structure provides more local information than when the whole image is represent as only a single vector as it is the case when e.g.\ the pool\_5 layer is used.

As can be observed in Figure \ref{fig:exp1}, where the proposed hybrid objective function was used to train the network for the image classification problem on CIFAR-10, increasing the amount of unlabeled samples resulted in increased classification performance. The highest test accuracy was achieved when 40\,000 unlabeled samples were used (yellow line). The same can be seen in Figure \ref{fig:ucf_2} and Table \ref{table:ucf_gmm}, where different amounts of unlabeled data were used to train the network for the action recognition problem on UCF-101. The highest test accuracy in this case was also achieved when the largest number of unlabeled samples was used during training. The effect of using a small number of M-step iterations in our algorithm are shown in Figure \ref{fig:exp2} where only a single iteration is used. It can be seen that the improvement in the training is not as stable as when more iterations are used as shown in e.g.\ Figure \ref{fig:exp1} and Figure \ref{fig:ucf_2} (both using 5-M step iterations).

In Table \ref{table:ucf_gmm_offline} we show that the performance of the network when the GMM part of it is trained offline surpasses the performance of the network trained using our semi-supervised online approach. As already mentioned, this is because the batch EM-algorithm used to train the GMM offline has access to all the data during each EM iteration, while our online algorithm only sees a single mini-batch at a time. We note that the batch EM cannot be used in all situations - all data might not be available immediately, or it might be impossible to apply the batch algorithm because of memory constraints. Our proposed training method does not have this problem.

We tried matching the performance of the network trained in a semi-supervised way with a network trained only with labeled data in an experiment shown in Table \ref{table:ucf_labeled_only}. We show that only 600 additional labeled samples were needed to match the performance gained from adding 8\,500 unlabeled samples.
We checked how the performance changes as a function of training time mini-batch size, shown in Table \ref{table:changing_mb_size}. It can be seen that the smallest mini-batch size resulted in the worst performance, as a single sample does not bring much information about the distribution being modeled. If the mini-batch is too large the performance again degrades, as it is often observed when training neural networks.

To conclude, we presented a novel semi-supervised Fisher vector network and showed can be applied on an image classification problem (on CIFAR-10) and an action recognition problem (on UCF-101). We believe that we have shown some promising results and further experiments should be performed to see how well the semi-supervised Fisher vector network would perform on the whole UCF-101 dataset when unlabeled data is added from a larger dataset, e.g.\ the YouTube-8M dataset \cite{abu2016youtube}.

{\small
\bibliographystyle{ieee}
\bibliography{bibliography}

\begin{thebibliography}{10}\itemsep=-1pt

\bibitem{abu2016youtube}
S.~Abu-El-Haija, N.~Kothari, J.~Lee, P.~Natsev, G.~Toderici, B.~Varadarajan,
  and S.~Vijayanarasimhan.
\newblock Youtube-8m: A large-scale video classification benchmark.
\newblock {\em arXiv preprint arXiv:1609.08675}, 2016.

\bibitem{arthur2007k}
D.~Arthur and S.~Vassilvitskii.
\newblock k-means++: The advantages of careful seeding.
\newblock In {\em Proceedings of the eighteenth annual ACM-SIAM symposium on
  Discrete algorithms}, pages 1027--1035. Society for Industrial and Applied
  Mathematics, 2007.

\bibitem{dempster1977maximum}
A.~P. Dempster, N.~M. Laird, and D.~B. Rubin.
\newblock Maximum likelihood from incomplete data via the {EM} algorithm.
\newblock {\em Journal of the royal statistical society. Series B
  (methodological)}, pages 1--38, 1977.

\bibitem{duchi2011adaptive}
J.~Duchi, E.~Hazan, and Y.~Singer.
\newblock Adaptive subgradient methods for online learning and stochastic
  optimization.
\newblock {\em Journal of Machine Learning Research}, 12(Jul):2121--2159, 2011.

\bibitem{krizhevsky2009learning}
A.~Krizhevsky.
\newblock Learning multiple layers of features from tiny images.
\newblock 2009.

\bibitem{neal1998view}
R.~M. Neal and G.~E. Hinton.
\newblock A view of the {EM} algorithm that justifies incremental, sparse, and
  other variants.
\newblock In {\em Learning in graphical models}, pages 355--368. Springer,
  1998.

\bibitem{palasek2016marmi}
P.~Palasek and I.~Patras.
\newblock Action recognition using convolutional restricted {B}oltzmann
  machines.
\newblock In {\em Proceedings of the 1st International Workshop on Multimedia
  Analysis and Retrieval for Multimodal Interaction}, MARMI '16, pages 3--8,
  New York, NY, USA, 2016. ACM.

\bibitem{palasek2017discriminative}
P.~Palasek and I.~Patras.
\newblock Discriminative convolutional fisher vector network for action
  recognition.
\newblock {\em arXiv preprint arXiv:1707.06119}, 2017.

\bibitem{scikit-learn}
F.~Pedregosa, G.~Varoquaux, A.~Gramfort, V.~Michel, B.~Thirion, O.~Grisel,
  M.~Blondel, P.~Prettenhofer, R.~Weiss, V.~Dubourg, J.~Vanderplas, A.~Passos,
  D.~Cournapeau, M.~Brucher, M.~Perrot, and E.~Duchesnay.
\newblock Scikit-learn: Machine learning in {P}ython.
\newblock {\em Journal of Machine Learning Research}, 12:2825--2830, 2011.

\bibitem{simonyan2014very}
K.~Simonyan and A.~Zisserman.
\newblock Very deep convolutional networks for large-scale image recognition.
\newblock In {\em International Conference on Learning Representations}, 2015.

\bibitem{tieleman2012lecture}
T.~Tieleman and G.~Hinton.
\newblock Lecture 6.5-rmsprop: Divide the gradient by a running average of its
  recent magnitude.
\newblock {\em COURSERA: Neural networks for machine learning}, 4(2), 2012.

\end{thebibliography}
}

\end{document}